\pdfoutput=1

\documentclass[11pt,a4paper,fleqn]{article}

\usepackage[margin=2.5cm]{geometry}
\usepackage[T1]{fontenc}
\usepackage[utf8]{inputenc}
\usepackage{lmodern}

\usepackage{amssymb,amsmath,amsfonts}
\usepackage{graphicx}
\usepackage{textcomp}
\usepackage{xcolor}
\usepackage{multirow}
\usepackage{float}
\usepackage{booktabs}
\usepackage{siunitx}
\usepackage{algorithm}
\usepackage{algpseudocode}
\usepackage{comment}
\usepackage{threeparttable}
\usepackage{capt-of}
\usepackage{placeins}

\usepackage{authblk}
\usepackage[authoryear,round]{natbib}

\usepackage{hyperref}
\usepackage[nameinlink]{cleveref}
\usepackage{caption}
\crefname{appendix}{Appendix}{Appendices}
\Crefname{appendix}{Appendix}{Appendices}

\title{Combining Statistical Features and Deep Encodings for\\
Rehearsal-Based Class-Incremental Time Series Classification}

\author[1,2]{Pablo García-Santaclara\thanks{Corresponding author: \texttt{pgarcia@alumnos.uvigo.es}}}
\author[2]{Bruno Fernández-Castro}
\author[1]{Rebeca P. Díaz-Redondo}

\affil[1]{atlanTTic -- ICLAB, Universidade de Vigo, Vigo 36310, Spain}
\affil[2]{Centro Tecnolóxico de Telecomunicacións de Galicia (GRADIANT), Vigo 36214, Spain}

\date{\today}

\begin{document}
\maketitle

\begin{abstract}
Many systems used in real-world environments require adding new categories and incorporating new information without forgetting what was previously learnt by the classification model. This is known as class-incremental continual learning, and in the case of multivariate time-series, is further complicated by the temporal structure of the data. In this paper, we present a novel approach for performing class incremental continual learning for the classification of multivariate time series data based upon the construction of a dual-stream feature extraction pipeline (using both deep temporal embedding features generated via a pre-trained frozen foundation model and application of statistical features). Evaluated on five benchmark datasets, the proposed system achieves competitive average accuracy across all datasets while maintaining low forgetting rates across all experimental configurations.
\end{abstract}


\bigskip


\section{Introduction}
\label{sec:intro}

Implementing a machine learning model in an online environment is challenging. It must be done in such a way that it can be adapted incrementally to new information, but without overwriting the past knowledge that the model already possesses. This phenomenon, known as catastrophic forgetting~\cite{kirkpatrick2017overcoming}, is the central problem in continual learning: when a model is updated on new data, optimization tends to overwrite the weights that contained previous knowledge. Despite the existence of significant research in the field of continual learning, most of it is focused on classification problems with image datasets~\cite{de2021continual}. In time series contexts, observations have temporal dependencies, and the relevant information is distributed throughout the entire sequence. Studies of continual learning in the context of time series are limited. In addition to the complexities inherent in any continual learning problem, time series add the difficulty of temporal dependencies and high intra-class variance, for example there are cases of samples belonging to the same class, can differ substantially across subjects, recording conditions, or sensor placement. The approaches commonly used for tabular data are not directly applicable to time series data.

Several strategies have been proposed to mitigate catastrophic forgetting. Regularisation-based methods impose constraints on the loss function to preserve prior knowledge, whilst architecture-based methods assign independent parameters to each task. In contrast, rehearsal methods store a subset of previous data, or generate synthetic approximations of it, and reuse this during training for new tasks. The last ones have usually shown the best results against catastrophic forgetting. A memory-efficient variant is using prototypes, which are compact representative vectors that summarize the distribution of each past class. 

Applying prototype-based rehearsal to time series data, however, introduces challenges that have not been previously addressed. Raw time series samples are multi-channel tensors with a temporal structure that cannot be directly represented as flat prototype vectors. Furthermore, time series datasets typically exhibit high intra-class variance, which destabilises prototype neighbourhoods. This motivates the need for a fixed feature representation that collapses the temporal dimension while preserving discriminative information. To balance effective representations with the goal of stability, we propose combining two complementary approaches: deep temporal embeddings derived from a frozen base model, which capture the overall temporal structure, and statistical features computed for each sample. To the best of our knowledge, this combination has not been explored in the context of continual class incremental learning for time series.

In this paper, we present a novel approach for class incremental learning for the classification of multivariate time-series data. Our system extends TRIL3~\cite{garcia2025overcoming}, a framework originally designed for tabular data. Concretely, the contributions of this paper are:
\begin{itemize}
    \item A dual-stream feature representation for class-incremental learning on multivariate time series, combining frozen foundation model embeddings with statistical and inter-channel correlation features.
    \item An extension of TRIL3 to multivariate time series, enabling class-incremental learning without hyperparameter tuning phases.

\end{itemize}

The resulting feature vector is cross-task stable, compatible with the prototype model, and sufficiently rich to enable accurate incremental classification. 

The rest of the article is structured as follows: ~\Cref{sec:related} reviews the literature on continuous learning for time series. ~\Cref{sec:methodology} describes the challenges and the proposed methodology in detail. ~\Cref{sec:experiments} presents the experimental setup, the results and the ablation analysis. ~\Cref{sec:conclusions} concludes the article.

\section{Related Work}
\label{sec:related}

There are several studies that focus on continual learning for time series. Some of the early work focused on specific application domains. Set in a clinical context, Kiyasseh et al.~\cite{kiyasseh2021clinical} demonstrated that a model continuously updated using cardiac signals from various diseases, institutions, and modalities is capable of maintaining accuracy against gradual and sudden changes in distribution. Meanwhile, Doshi and Yilmaz~\cite{doshi2022rethinking} approached video anomaly detection as a time series problem, combining pre-trained spatio-temporal feature extractors with a kNN-based memory module and an RNN to capture temporal dependencies in anomaly detection. Schillaci et al.~\cite{schillaci2021prediction} addressed distribution shifts in greenhouse monitoring through a prediction error-driven memory consolidation mechanism, retaining samples based on their expected importance for learning progress. More recent contributions have extended this to other operational domains: Xiao et al.~\cite{xiao2022streamingtrafficflowprediction} combined online updating with reinforcement learning for real-time traffic flow modelling, triggering updates only when sustained prediction error exceeds a dynamic threshold, while Aslam et al.~\cite{Aslam2025CEL} and Wang et al.~\cite{WANG2025122614} addressed conceptual drift in disease forecasting and urban drainage modelling respectively, through domain adaptation and experience replay. These contributions illustrate different scenarios where continual learning for time series is useful, as also surveyed by Ao and Fayek~\cite{ao2023continual}, who provide a systematic review of deep learning applications for sensor time series data in the context of continual learning.

The lack of benchmarks in this field was highlighted by Qiao et al.~\cite{qiao2024class}, who proposed a standardised benchmark for class-incremental time series learning that enables systematic comparison across methods and datasets, and used it to evaluate various continuous learning approaches in this context. Their work establishes the evaluation framework adopted in this paper.

Furthermore, although relevant to our approach, the field of time series representation has recently been expanded with the emergence of large pre-trained foundation models. In recent years, some foundation models dedicated to time series data have emerged, trained on large amounts of data. Models such as TimesFM~\cite{das2024decoderonlyfoundationmodeltimeseries}, MOMENT~\cite{goswami2024moment}, and Chronos~\cite{ansari2024chronoslearninglanguagetime}  demonstrate that transformers pre-trained on large and diverse time series corpora can produce representations that transfer across domains and tasks without fine-tuning. TimesFM is Google Research’s pretrained time-series foundation model, designed as a decoder-only model, it is used exclusively to forecast future values based on past sequences. Chronos is a pretrained time-series foundation model that reframes forecasting as a language modeling problem instead of direct numeric regression, as TimesFM it can only be used for forecasting tasks. MOMENT, by contrast, is pre-trained under a multi-task objective that includes reconstruction and embedding, making it applicable to classification and representation tasks beyond forecasting. More recently, models such as TiCT~\cite{yeh2025tictsyntheticallypretrainedfoundation} have explored in-context learning for time series classification. The most directly comparable prior work is PTMs-TSCIL~\cite{wu2025ptmstscilpretrainedmodelsbased}, which also uses MOMENT as a backbone for class-incremental time series learning. Unlike our approach, PTMs-TSCIL keeps the backbone frozen while incrementally tuning a shared adapter per task, using knowledge distillation to regularize adapter updates and prevent over-adaptation. To address the feature drift introduced by adapter tuning, it introduces a Drift Compensation Network. While PTMs-TSCIL demonstrates good results, its pipeline introduces significant architectural complexity: adapter tuning, knowledge distillation, and a two-stage DCN training procedure are used.

The studies analysed above indicate that there exists growing interest in the time series continual learning field. Most contributions, such as those by Kiyasseh et al.~\cite{kiyasseh2021clinical}, Schillaci et al.~\cite{schillaci2021prediction} and Wang et al.~\cite{WANG2025122614}, focus on specific application domains and do not address the general class-incremental setting. TSCIL~\cite{qiao2024class} formalises this as a general benchmark, evaluating methods that are adaptations of existing continual learning approaches, and it reserves a subset of classes exclusively for hyperparameter tuning, meaning the deployed model never learns the full class set. PTMs-TSCIL~\cite{wu2025ptmstscilpretrainedmodelsbased} is the most directly related work, introducing pre-trained models into the TSCIL setting through adapter tuning and a drift compensation network. Our work proposes a fully frozen encoder whose representations are complemented by statistical and inter-channel correlation features, producing a more complete representation than deep embeddings alone. Combined with prototype-based rehearsal, this approach requires no adapter tuning, no knowledge distillation, and no drift compensation mechanism, while remaining online and being able of covering the full class set without any tuning phase.

\section{Methodology}
\label{sec:methodology}

In this section, we first discuss the different challenges involved in applying continual learning to a time series problem in~\Cref{subsec:Challenges}. We then overview the prototype classification framework, which serves as a basis for this work. Finally, we explore the design choices for the statistical representation and foundational model embedding of the samples in~\Cref{subsec:feature_rep}. 

\subsection{Challenges}
\label{subsec:Challenges}

The first challenge that arises is that raw time-series samples cannot be used with a prototype model. Prototypes are feature vectors, with no mechanism for handling cross-channel relationships or capturing temporal dependencies. In contrast, time-series samples are three-dimensional tensors composed of $N$ sequences (or samples) of $C$ channels (or features) and $T$ time steps. Therefore, a vector representation is needed that can be used to generate the prototypes, and the same representation must also be used for the classification model.

The second challenge is that time series problems can involve high intra-class variance. In typical time series datasets, data within the same class, such as the execution of a specific gesture, can vary greatly depending on the person doing it. This variability complicates the performance of any classification model, but it also makes prototype neighbourhoods less stable.

To address these challenges, we introduce the following adaptations to 
TRIL3:

\begin{enumerate}
    \item \textbf{Foundational model embeddings}: We incorporate a pre-trained time series foundation model to extract deep temporal representations, producing a fixed-length embedding per sample that captures global temporal structure.  As the encoder is pre-trained on a large corpus of time series, it generalizes across domains without requiring any retraining when applied to new data.
    \item \textbf{Statistical features}: We complement the embeddings with a set of interpretable time and frequency domain features computed per channel, together with inter-channel Pearson correlation coefficients that capture coordinated multi-channel activity.
    \item \textbf{Combined feature representation}: The two streams are concatenated into a unified feature vector, which is then used as input to both the prototype model and the classifier model.
\end{enumerate}

Raw input samples are first processed through the dual-stream feature extraction pipeline, which produces a unified feature vector consumed by both the prototype model and the classifier. The prototype model generates synthetic rehearsal samples from past classes, which are combined with real data from the current task to train the classifier incrementally.

\subsection{TRIL3 Overview}
\label{subsec:background}

This approach is based partly on the TRIL3~\cite{garcia2025overcoming} framework, which tackles the problem of catastrophic forgetting for tabular data using a generative rehearsal approach. The main contribution of TRIL3 is the use of an online prototype model to retain past knowledge. Although it is agnostic with regard to both the prototype model and the classifier model, in the original paper the prototype model used is XuILVQ~\cite{gonzalez2024decentralized} and the classifier model is a Deep Neural Decision Forest~\cite{kontschieder2015deep}. TRIL3 works by updating the prototype model in real time with real data, storing representative centroids which are then used to generate synthetic data; this, together with the real samples, is used to train the classification model.

\begin{figure}[H]
  \centering
    \includegraphics[width=.78\textwidth]{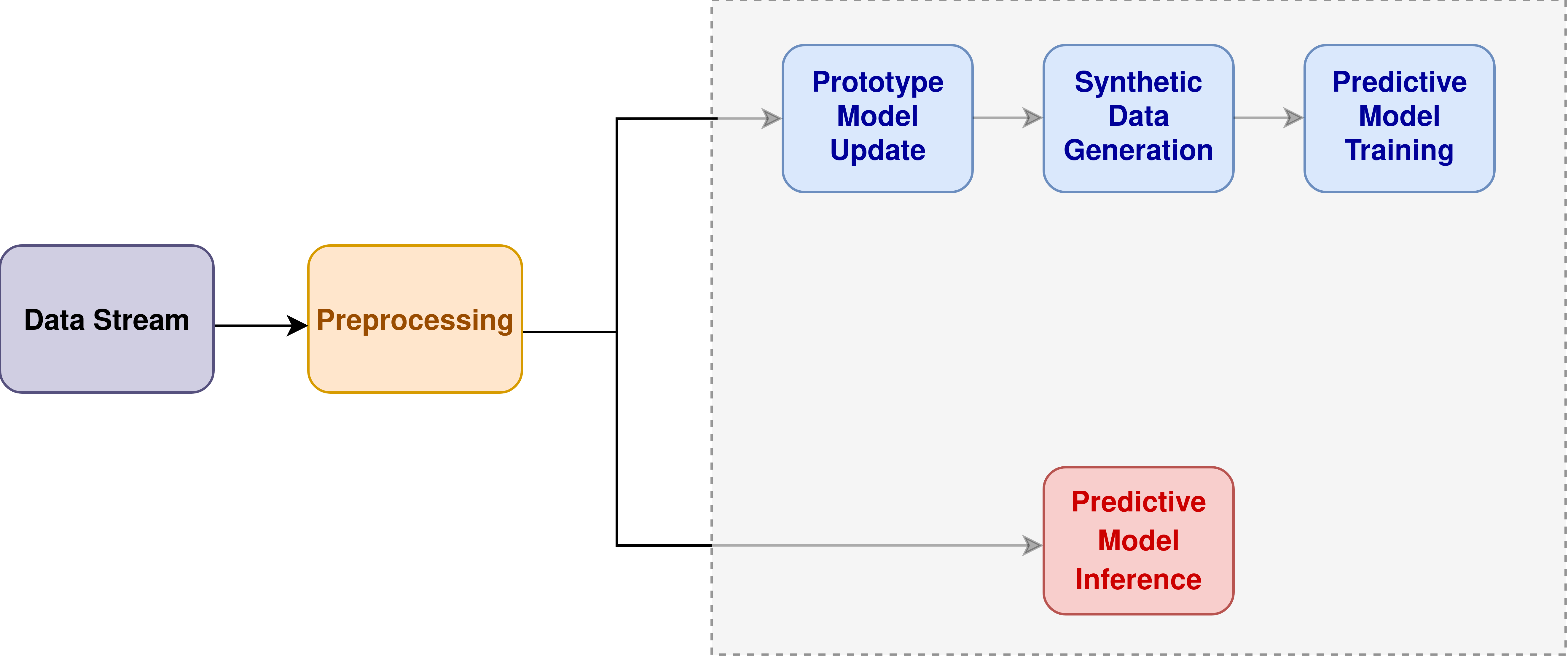}
  \caption{Simplified diagram of the TRIL3 framework.}
  \label{fig:tril3-workflow}
\end{figure}

~\autoref{fig:tril3-workflow} shows a simplified diagram of how the framework works. Initially, the prototype model is initialised with an empty set and the classifier model is initialised with random weights. The input data is normalised using a scaler that is also trained online. Next, labelled samples are used to update the prototype set, thereby creating new prototypes or updating existing ones. Synthetic samples are extracted pseudo-randomly from the prototype set and mixed with real data to form a balanced training batch. The classification model is trained on the batch of real and synthetic data, although training only occurs when the prototype set has grown, thereby avoiding unnecessary updates and reducing the risk of overfitting. If the input data is unlabelled, the set of real samples is used for inference with the classifier model. However, TRIL3 was originally designed for the classification of tabular data, and its application to time-series data requires several modifications to its architecture so that it can handle the temporal structure of the input data.

\subsection{Feature Representation}
\label{subsec:feature_rep}

The proposed feature extraction process processes each raw time series sample independently through two parallel pipelines: a pre-trained foundational model stream for time series that requires no fine-tuning, and a statistical features stream that calculates time, frequency, and correlation descriptors. The outputs from both pipelines are concatenated into a unified feature vector.

\subsubsection{Foundational Model Stream}

The first stream consists of a pre-trained model that generates embeddings from the input data. The weights of this model remain frozen, as it is never retrained or fine-tuned. This prevents catastrophic forgetting at the feature extraction level; if the encoder were retrained, the representations of past classes would shift, invalidating the stored prototypes. The goal of using a foundational model is to map samples spanning multiple channels and time steps to a single point, thereby collapsing the temporal dimension and eliminating the sequential dependencies inherent in time series data. The resulting embedding $\mathbf{e} \in \mathbb{R}^{1024}$ serves as a compact representation that can be directly consumed by both the prototype model and the classifier.

The encoder chosen for the implementation is MOMENT~\cite{goswami2024moment}, a foundational transformer-based model specialized in time series data. MOMENT serves as a general purpose model because it is pretrained on a large and diverse corpus of time series datasets spanning multiple domains, which means it does not require fine-tuning to perform well. Unlike models such as Chronos~\cite{ansari2024chronoslearninglanguagetime} and TimesFM~\cite{das2024decoderonlyfoundationmodeltimeseries}, which are 
designed exclusively for forecasting, MOMENT is pretrained under a multi-task objective that includes embedding, making it directly suited for our purpose. There are three MOMENT models: large, small and base. The size of the output vector depends on the model used, for our implementation the large model was chosen. Given an input sample of shape $(N, C, T)$, where $N$ is the number of samples, $C$ is the number of channels, and $T$ is the number of time steps, MOMENT produces a fixed-length embedding of dimension $1024$ per sample via mean pooling over the temporal dimension. 

\subsubsection{Statistical Features Stream}

The second stream extracts a set of statistical features computed 
independently for each time series sample, with no operation that crosses 
sample boundaries. The selected features were inspired by established time series feature extraction frameworks~\cite{BARANDAS2020100456, CHRIST201872}, which organize descriptors across time and frequency domains. However, rather than extracting the full set of available features, we selected a compact subset of 14 features per channel trying to balance representational richness with efficiency. The features extracted per channel are: mean, standard deviation, minimum, maximum, range, mean absolute deviation (MAD), root mean square (RMS), skewness, kurtosis, and zero-crossing rate in the time domain; and spectral energy, dominant frequency, spectral centroid, and spectral spread in the frequency domain. The definition of the features can be found at Appendix~\ref{app:features}. Time-domain features capture amplitude and energy (mean, std, min, max, range, MAD, RMS), distributional shape (skewness, kurtosis), and signal dynamics (zero-crossing rate). Frequency-domain features characterise the spectral content and energy distribution (spectral energy, dominant frequency, spectral centroid, spectral spread). Together, these features describe each channel's signal from complementary perspectives without requiring any learned parameters, making them inherently stable across incremental tasks. The result is a feature vector of dimension $C \times 14$ per sample. The features are calculated from the original, unnormalized signal, since any normalization would destroy the signal's absolute scale, making features such as mean, RMS, and spectral energy meaningless. The complete feature vector of a sample $\mathbf{x} \in \mathbb{R}^{C \times T}$ is obtained by applying a channel-wise feature extraction function, which computes the 14 features described above independently for each channel and concatenates the results, yielding $\mathbf{h} \in \mathbb{R}^{C \times 14}$.

In addition to the statistics mentioned above, to maximize the information provided by the signal, we also calculate the correlation coefficients between the channels. We use the Pearson correlation matrix for each sample. For a time-series  $\mathbf{x} \in \mathbb{R}^{C \times T}$, the Pearson correlation matrix $\mathbf{M} \in \mathbb{R}^{C \times C}$ is computed, where each entry $M_{ij}$ is the linear co-variation between channels $i$ and $j$. Since $\mathbf{M}$ is symmetric with unit diagonal, only the upper triangular entries are retained, yielding a vector $\mathbf{r} \in \mathbb{R}^{C(C-1)/2}$ of unique correlation coefficients per sample. The inter-channel correlation is very useful in datasets where class identity is defined by coordinated multi-channel activity.

Like the foundation model stream, the statistical features are computed by a fixed deterministic function with no learnable parameters, giving them the same stability guarantee by construction.

\subsubsection{Combined Representation}

The outputs of both streams are concatenated to form the final vector:

\begin{equation}
    \mathbf{f} = [\mathbf{e} \| \mathbf{h} \| \mathbf{r}]
\end{equation}

\noindent where $\mathbf{e} \in \mathbb{R}^{1024}$ is the MOMENT embedding, 
$\mathbf{h} \in \mathbb{R}^{C \times 14}$ is the per-channel statistical feature vector, and $\mathbf{r} \in \mathbb{R}^{C(C-1)/2}$ is the cross-channel correlation vector, yielding a combined representation of dimension $1024 + C \times 14 + C(C-1)/2$.

\begin{figure}[H]
    \centering
    \includegraphics[width=0.8\textwidth]{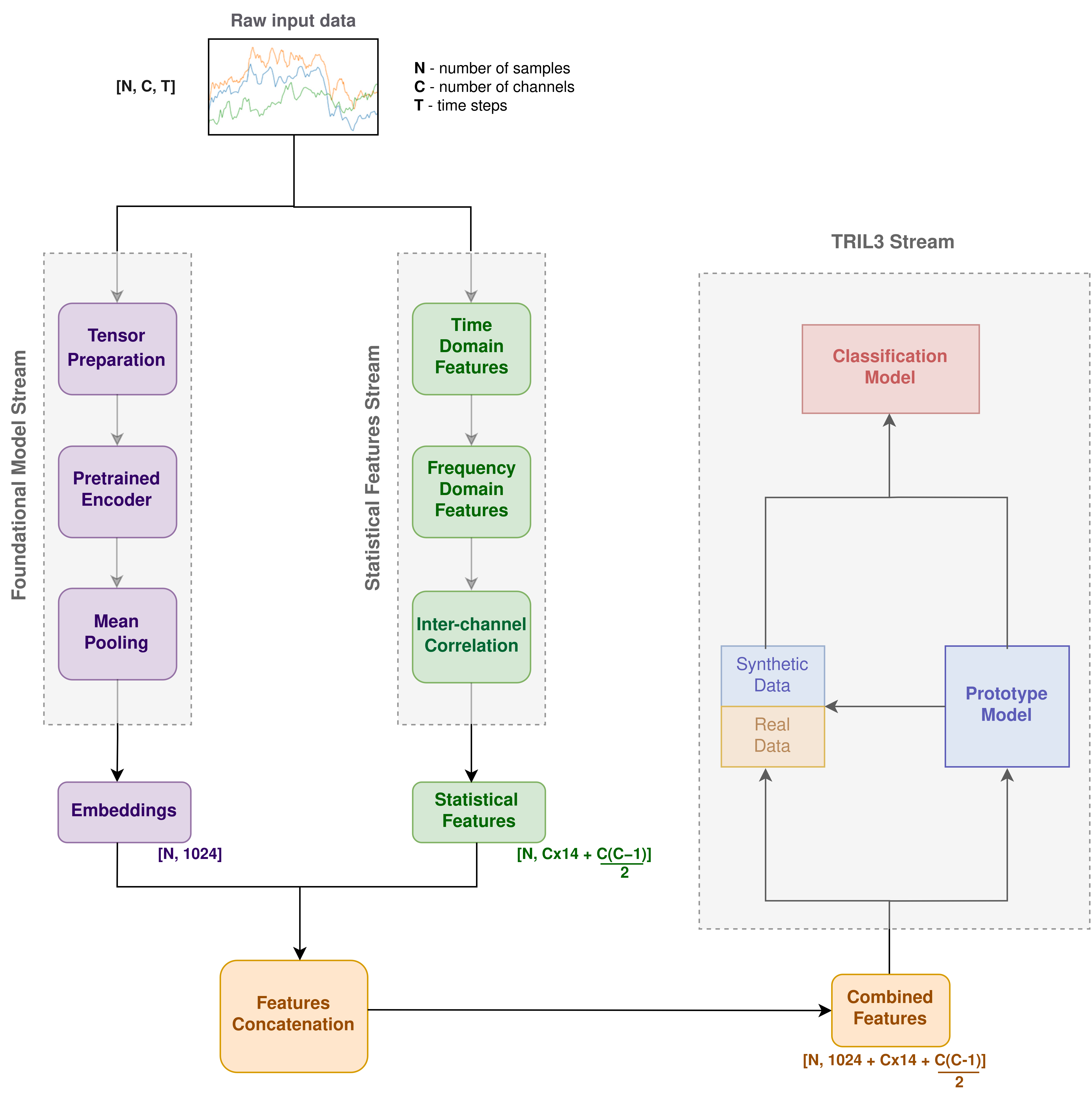}
    \caption{Diagram of the dual-stream pipeline. Embeddings and statistical features are extracted in parallel and concatenated into a unified feature vector for the prototype model and classifier.}
    \label{fig:whole-arch}
\end{figure}

\autoref{fig:whole-arch} shows the dual-stream pipeline in detail. The foundational model stream and the statistical features stream process the raw input independently, producing $\mathbf{e}$, $\mathbf{h}$, and $\mathbf{r}$ respectively, which are then concatenated into the final feature vector $\mathbf{f}$. Because both streams are individually stable, their concatenation inherits this stability.
\section{Experiments / Validation and Results}
\label{sec:experiments}

The experiments conducted to evaluate the proposed framework are similar to those carried out by the TSCIL benchmark~\cite{qiao2024class}, a study by Qiao et al. that evaluated several continual learning techniques in an incremental learning class setting on time series data. TSCIL acts as a standardized framework, also providing a set of pre-processed datasets with defined training and test sets, which facilitates reproducibility. Therefore, for our experiments, we directly use its datasets, pre-processing steps, and task configurations. In addition to TSCIL, we include PTMs-TSCIL~\cite{wu2025ptmstscilpretrainedmodelsbased} as a second point of reference which evaluates on four of the same five datasets, GRABMyo is not included in their evaluation. Some methodological differences limit direct numerical comparison: PTMs-TSCIL applies PCA to reduce DSA from 45 to 15 channels before feature extraction, uses a hyperparameter validation stream following the TSCIL tuning protocol, and averages results over 3 runs rather than 5. 

In our evaluation, we analyse whether the proposed dual-stream feature representation, combined with the TRIL3 process, performs competitively against the methods described in the TSCIL benchmark. Secondly, through an ablation study, we quantify the individual contribution of each stream to the final result. Beyond classification performance, the proposed system offers two additional properties relevant to real-world deployment. First, since no raw training samples are stored at any point, only synthetic prototypes generated from learned centroids, the system does not retain identifiable data from past tasks. Second, the entire pipeline operates in an online way: both the prototype model and the classification model are updated incrementally as each sample arrives, with no requirement for batch access to task data or offline retraining phases.

\subsection{Experimental Setup}
\label{subsec:setup}

To enable a direct comparison with the TSCIL benchmark, we adopt the same five datasets used in that work: DSA~\cite{ismidewi}, GRABMyo~\cite{pradhan}, WISDM~\cite{wisdm_smartphone_and_smartwatch_activity_and_biometrics_dataset__507}, UCIHAR~\cite{altunkerem}, and UWave~\cite{liujiayang}. They belong to two application domains: human activity recognition and gesture recognition. DSA captures motion sensor data across 19 daily sports activities performed by 8 volunteers. GRABMyo is a large-scale surface electromyography database for hand gesture recognition, comprising 16 gestures performed by 43 participants across 28 channels. WISDM is a sensor-based human activity recognition dataset covering 18 activities from 51 subjects, using the phone accelerometer modality with a window size of 200 timesteps at 20 Hz. UCI-HAR contains inertial sensor sequences from smartphones recorded at 50 Hz while 30 volunteers performed 6 daily activities. UWave comprises accelerometer recordings from 8 subjects performing 8 gesture patterns. The train/test splits are produced using the original preprocessing scripts provided by the authors of TSCIL. Table~\ref{tab:datasets} summarizes the key characteristics of each dataset.

\begin{table}[H]
\centering
\caption{Overview of the benchmark datasets and task configurations.}
\label{tab:datasets}
\begin{tabular}{lcccc}
\toprule
\textbf{Dataset} & \textbf{Shape} $(C \times T)$ & \textbf{Classes} & \textbf{Exp. classes} & \textbf{Tasks} \\
\midrule
UCI-HAR  & $9 \times 128$  & 6  & 6  & 3 \\
UWave    & $3 \times 315$  & 8  & 8  & 4 \\
DSA      & $45 \times 125$ & 18 & 12 & 6 \\
GRABMyo  & $28 \times 128$ & 16 & 12 & 5 \\
WISDM    & $3 \times 200$  & 18 & 12 & 6 \\
\bottomrule
\end{tabular}
\end{table}

Following the evaluation protocol of TSCIL, three of the five datasets (DSA, GRABMyo, and WISDM) are evaluated in a subset configuration in which only 12 out of their full set of classes are used. In the original paper, this reduction is intended to reserve classes for preliminary tasks in which the model is hyperparameterised. In our proposal, there are no tasks requiring fine-tuning; a single fixed set of hyperparameters is applied to all datasets and all experimental conditions. The same subsets are retained to ensure that class distributions remain comparable. UCIHAR and UWave are evaluated using their full set of classes on both systems.

\subsection{Evaluation Protocol}
\label{subsec:protocol}

The evaluation is carried out using an incremental class learning protocol; classes are introduced sequentially into independent tasks, and the model has no access to data from previous tasks, either during training or during inference. As with TSCIL, each task consists of two new classes, which are randomly selected for each task in each run.

\subsubsection{Hyperparameter Tuning}

Our approach is methodologically different in the way it handles hyperparameter tuning. For datasets with a large number of tasks, TSCIL partitions the full class set into two separate streams: a \textit{validation stream} of 3 tasks used exclusively for hyperparameter search, and an \textit{experiment stream} used for training and evaluation. As a consequence, the classes assigned to the validation stream are never learned by the final model; they are consumed entirely by the tuning process. This means that in practice, the deployed model has knowledge of only a subset of the available classes in the domain, which is a significant limitation in any real-world application where complete class coverage is required. For UCI-HAR and UWave, which have fewer tasks, TSCIL instead uses a per-task train/validation split for tuning. Our system imposes no such constraint. A single fixed set of hyperparameters is used across all datasets with no validation stream and no tuning phase. This scenario is closer to reality: the model is trained on all available classes without reserving any for tuning, and the same configuration generalizes across datasets of varying complexity, channel count, and sequence length.

\subsubsection{Metrics}

Performance is measured using two standard continual learning metrics,
following the same definitions as TSCIL. Let
$a_{i,j}$ denote the accuracy evaluated on the test set of task $j$
after the model has been trained on task $i$, where $j \leq i$.

\begin{itemize}
    \item \textbf{Average accuracy} ($\mathcal{A}_i$): defined as
    $\mathcal{A}_i = \frac{1}{i} \sum_{j=1}^{i} a_{i,j}$, the mean
    per-task accuracy after learning task $i$.  $\mathcal{A}_T$  denotes the final value after all T tasks, used as the primary performance metric.

    \item \textbf{Average forgetting} ($\mathcal{F}_T$): defined as
    $\mathcal{F}_T = \frac{1}{T-1} \sum_{j=1}^{T-1} f_{T,j}$, where
    $f_{T,j} = \max_{k \in \{1,\dots,T-1\}} (a_{k,j}) - a_{T,j}$
    is the drop from the peak accuracy ever achieved on task $j$ to
    its accuracy at the end of training. Lower values indicate less
    forgetting.
\end{itemize}

\subsection{Implementation Details}
\label{subsec:implementation}

Table~\ref{tab:hyperparams} summarizes the hyperparameters used for each component. A single fixed configuration is used across all datasets and experimental conditions. Prototypes generated by XuILVQ are used as synthetic samples. Two memory configurations are evaluated: \textbf{10\%} and \textbf{20\%} of real training samples per batch. All experiments are run with 5 different random seeds.

\begin{table}[H]
\centering
\caption{Hyperparameter configuration for each component of the proposed system.}
\label{tab:hyperparams}
\begin{tabular}{clc}
\toprule
\textbf{Component} & \textbf{Parameter} & \textbf{Value} \\
\midrule
& Number of trees       & 50 \\
& Tree depth            & 12 \\
\textbf{DNDF}
& Feature rate          & 1.0 \\
& Learning rate         & 0.01 \\
& Batch size            & 64 \\
\midrule
& Model variant         & MOMENT-1-large \\
& Task                  & Embedding \\
\textbf{MOMENT}
& Output dimension      & 1024 \\
& Temporal reduction    & Mean pooling \\
& Batch size            & 128 \\
\midrule
& $\alpha_{\text{winner}}$     & 0.9  \\
& $\alpha_{\text{runner-up}}$  & 0.1  \\
\textbf{XuILVQ}
& Age threshold                & 600 \\
& $\gamma$                     & 800 \\
& Target prototypes per class  & 5 \\
& Initial prototypes per class & 50 \\
\bottomrule
\end{tabular}
\end{table}

\subsection{Ablation Study: Encoding Configurations}
\label{subsec:ablation}

To assess the contribution of each component of the dual-stream feature extraction pipeline, we evaluate three encoding configurations:

\begin{itemize}
    \item \textbf{Foundational Model Embeddings}: only the foundational model embedding $\mathbf{e}$ is used as input.
    \item \textbf{Statistical Features}: only the statistical feature vector $[\mathbf{h} \| \mathbf{r}]$ is used as input.
    \item \textbf{Full Representation}: the complete feature vector $\mathbf{f} = [\mathbf{e} \| \mathbf{h} \| \mathbf{r}]$ is used as input.
\end{itemize}

This ablation isolates the relative contribution of the embeddings and the statistical features, and quantifies the benefit of using both.

\subsection{Results}
\label{subsec:results}

\begin{figure}[H]
    \centering
    \includegraphics[width=0.92\textwidth]{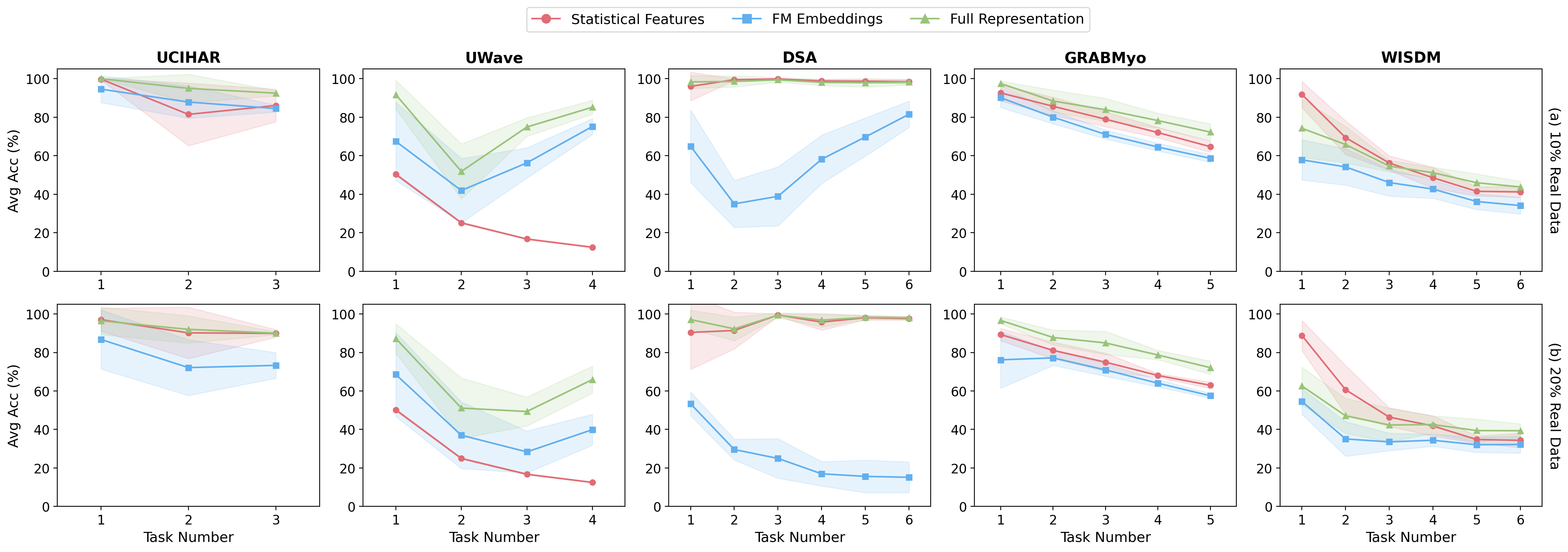}
    \caption{Evolution of average accuracy $\mathcal{A}_i$ using (a) 10\% Real Data and (b) 20\% Real Data ratios.}
    \label{fig:curves}
\end{figure}

\autoref{tab:results_comparison} reports the comparison between our system, the TSCIL baseline and the PTMs-TSCIL approach across all five datasets. TSCIL results are reported for the best-performing regularization and replay method respectively, under both BatchNorm and LayerNorm normalization configurations.~\autoref{fig:curves} allows us to analyze the evolution of the average accuracy of our proposal for each dataset across the tasks.

\begin{table}[h]
\centering
\caption{Average accuracy $\mathcal{A}_T$ and average forgetting $\mathcal{F}_T$ comparison across the five datasets.}
\label{tab:results_comparison}
\resizebox{\textwidth}{!}{%
\begin{tabular}{ll cccc c cccccc}
\toprule
& & \multicolumn{4}{c}{\textbf{TSCIL}} & \textbf{PTMs-TSCIL} & \multicolumn{6}{c}{\textbf{Ours}} \\
\cmidrule(lr){3-6} \cmidrule(lr){7-7} \cmidrule(lr){8-13}
& & \multicolumn{2}{c}{Best Reg.} & \multicolumn{2}{c}{Best Replay}
  & 
  & \multicolumn{2}{c}{FM Embeddings} & \multicolumn{2}{c}{Statistical Features} & \multicolumn{2}{c}{Full Representation} \\
\cmidrule(lr){3-4} \cmidrule(lr){5-6} \cmidrule(lr){8-9} \cmidrule(lr){10-11} \cmidrule(lr){12-13}
\textbf{Dataset} & \textbf{Metric}
  & BN & LN & BN & LN
  & 
  & 10\% & 20\% & 10\% & 20\% & 10\% & 20\% \\
\midrule
\multirow{2}{*}{UCI-HAR}
  & $\mathcal{A}_T$ & $51.5_{\pm14.1}$ & $77.9_{\pm15}$   & $89.1_{\pm6.1}$  & $90.9_{\pm1.7}$ & $88.4_{\pm10.3}$ & $84.5_{\pm2.0}$ & $73.3_{\pm7.4}$  & $86.1_{\pm9.4}$  & $89.9_{\pm2.2}$  & $\textbf{92.5}_{\pm1.6}$ & $90.0_{\pm0.9}$ \\
  & $\mathcal{F}_T$ & $62.0_{\pm16.3}$ & $9.3_{\pm6.6}$   & $13.2_{\pm11.4}$ & $8.5_{\pm5.7}$  & $10.5_{\pm25.3}$ & $9.3_{\pm7.7}$  & $20.0_{\pm18.9}$ & $15.6_{\pm14.4}$ & $\textbf{2.5}_{\pm6.3}$  & $3.9_{\pm5.8}$           & $7.4_{\pm6.1}$  \\
\midrule
\multirow{2}{*}{UWave}
  & $\mathcal{A}_T$ & $62.7_{\pm11.6}$ & $58.1_{\pm18.3}$ & $83.2_{\pm3.5}$  & $85.2_{\pm2.2}$ & $85.1_{\pm9.7}$  & $75.2_{\pm4.6}$ & $40.0_{\pm9.0}$  & $12.5_{\pm0.2}$  & $12.5_{\pm0.2}$  & $\textbf{85.3}_{\pm4.0}$ & $65.9_{\pm7.9}$ \\
  & $\mathcal{F}_T$ & $32.9_{\pm12.0}$ & $35.4_{\pm25.6}$ & $20.3_{\pm4.5}$  & $16.6_{\pm2.4}$ & $11.1_{\pm15.7}$ & $\textbf{2.5}_{\pm11.8}$ & $27.7_{\pm13.1}$ & $13.6_{\pm14.0}$ & $13.3_{\pm13.8}$ & $9.6_{\pm6.3}$  & $24.8_{\pm16.2}$\\
\midrule
\multirow{2}{*}{DSA}
  & $\mathcal{A}_T$ & $31.3_{\pm6.4}$  & $35.9_{\pm5.6}$  & $96.9_{\pm2.2}$  & $98_{\pm1}$     & $96.8_{\pm6.5}$ & $81.5_{\pm7.6}$ & $15.2_{\pm8.9}$  & $\textbf{98.4}_{\pm1.3}$ & $97.6_{\pm0.9}$  & $98.2_{\pm1.7}$ & $98.0_{\pm0.9}$ \\
  & $\mathcal{F}_T$ & $60.0_{\pm11.5}$ & $53.0_{\pm9.1}$  & $3.7_{\pm2.7}$   & $2.2_{\pm1.2}$  & $2.8_{\pm1.7}$  & $\textbf{-2.0}_{\pm8.2}$ & $24.2_{\pm19.4}$ & $-1.3_{\pm1.2}$  & $2.0_{\pm0.7}$   & $1.4_{\pm2.2}$  & $1.7_{\pm0.6}$  \\
\midrule
\multirow{2}{*}{GRABMyo}
  & $\mathcal{A}_T$ & $20.7_{\pm5.2}$  & $20.8_{\pm7.1}$  & $55.9_{\pm5.1}$  & $64.5_{\pm3.5}$ & --                        & $58.7_{\pm2.2}$ & $57.5_{\pm1.4}$  & $64.7_{\pm3.1}$  & $62.9_{\pm1.8}$  & $\textbf{72.3}_{\pm4.8}$ & $72.1_{\pm3.8}$ \\
  & $\mathcal{F}_T$ & $49.8_{\pm25.3}$ & $21.9_{\pm10.4}$ & $31.1_{\pm4.9}$  & $36.6_{\pm3.4}$ & --                        & $20.8_{\pm5.0}$ & $20.0_{\pm5.2}$  & $21.3_{\pm4.9}$  & $\textbf{19.7}_{\pm1.1}$ & $23.4_{\pm5.4}$ & $22.5_{\pm5.8}$ \\
\midrule
\multirow{2}{*}{WISDM}
  & $\mathcal{A}_T$ & $16.4_{\pm6.6}$  & $21.4_{\pm6.6}$  & $51.6_{\pm13.7}$ & $\textbf{68.0}_{\pm7.6}$ & $36.1_{\pm8.6}$ & $34.1_{\pm4.8}$ & $32.2_{\pm5.0}$  & $41.2_{\pm3.1}$  & $34.4_{\pm4.2}$  & $43.8_{\pm3.3}$ & $39.3_{\pm4.0}$ \\
  & $\mathcal{F}_T$ & $63.8_{\pm44.0}$ & $41.2_{\pm21.5}$ & $27.9_{\pm12.4}$ & $31.9_{\pm7.9}$ & $30.7_{\pm5.9}$ & $\textbf{9.7}_{\pm5.6}$  & $10.2_{\pm5.9}$  & $26.0_{\pm8.4}$  & $28.5_{\pm13.3}$ & $13.7_{\pm6.9}$ & $15.2_{\pm6.5}$ \\
\bottomrule
\end{tabular}}
\end{table}

Table \ref{tab:results_comparison} shows that our proposal
outperforms the best TSCIL configurations on four of the five datasets in terms of average accuracy, while also improving results on catastrophic forgetting across all datasets. Also, despite the advantages in the methodology comparison described in the experimental setup, our system achieves superior results on every shared dataset. On UCI-HAR, the full representation with 10\% memory achieves 92.5\% of $\mathcal{A}_T$, surpassing both TSCIL's best result (90.9\%) and PTMs-TSCIL (88.4\%). It should be emphasized that these results are remarkable since, in our case, there is no prior hyperparameterization phase. On GRABMyo, the full representation achieves 72.3\% of $\mathcal{A}_T$, significantly surpassing TSCIL’s best result (64.5\%). In UWave, the 10\% full representation almost matches but slighly improves TSCIL and PTMs-TSCIL best reproduction results (85.3\%). On DSA, the statistical features stream achieves 98.4\% $\mathcal{A}_T$,  This near-perfect  performance is consistent with the low intra-class variance of DSA. On WISDM, TSCIL’s best result is significantly better (68.0\%). The WISDM dataset presents a particular challenge for frozen-encoder approaches, as both our system and PTMs-TSCIL struggle relative to TSCIL's best replay configuration, a challenge that deserves further investigation.

In terms of forgetting, our approach proves to be much more robust than the methods used by TSCIL and PTMs-TSCIL; the prototype system is more stable. In some cases, such as DSA, forgetting is negative, which likely means that during each task there were not enough samples for the classifier model to converge, so it continued to learn and improve thanks to the prototypes. As with TSCIL, the datasets that exhibited the highest rates of forgetting were GRABMyo and WISDM, since they are by far the largest datasets and also contain a large number of tasks. In contrast, for UCI-HAR and DSA, there is almost no forgetting across tasks.

Anomaly arises in the foundation model embeddings stream at 20\% memory, which collapses to 15.2\% $\mathcal{A}_T$ with 24.2\% $\mathcal{F}_T$, compared to strong performance at 10\%. Just as we explained earlier, negative forgetting in this dataset was due to the fact that there were many tasks and few samples per task; this same factor likely causes it to break down when the number of prototypes per sample is lower, in a dataset where embeddings provide less information than statistical features

The ablation study allows us to understand the contributions of the streams in each dataset. In some cases, such as DSA, it is the statistical features that provide the most important information for the full representation, this is substantially explained because, since this dataset has by far the most channels, the statistical data provides a great deal of information. In other datasets like UWave, the factors that improve the final results come from the embeddings, which is a dataset with only three channels, and therefore provides less information about the statistical features. However, in almost all cases, combining both results in a more comprehensive representation, which yields the best classification performance. Increasing the memory ratio from 10\% to 20\% does not consistently improve performance using the full representation. On UCI-HAR, DSA, GRABMyo and WISDM, the differences between 10\% and 20\% are small and within the standard deviation of the results. With UWave, there is indeed a significant difference between the two ratios, mainly due to the high level of forgetting associated with the 20\% 
\section{Conclusions}
\label{sec:conclusions}

This paper presents a novel approach for class-incremental continual 
learning on multivariate time series data, extending a prototype-based rehearsal framework with a dual-stream feature extraction pipeline that combines deep temporal embeddings from a frozen foundation model with statistical features. The resulting representation is stable across tasks, requires no hyperparameter tuning phase, and enables the model to be trained on the full class set under a single fixed configuration across all datasets.

The proposed system outperforms the best TSCIL configurations on four of the five datasets and PTMs-TSCIL on all comparable datasets in terms of average accuracy, while achieving substantially lower forgetting across all datasets. The ablation study reveals that the two streams play complementary roles that depend on the characteristics of each dataset. On high-channel datasets, the statistical representation is sufficient on its own. On low-channel datasets, the foundation model embeddings are the primary source of discriminative 
information. 

Beyond its performance in the classification task, the proposed system offers practical advantages for real-world systems. Since no raw training samples are ever retained, the system does not store data from previous tasks, which is important in many fields. Furthermore, its fully online operation eliminates any need for a prior training phase.

Several directions remain open for future work. Extending the evaluation 
to additional datasets beyond the TSCIL benchmark, focused on other areas, such as data from industrial processes. Another point to investigate is the performance difference observed in WISDM, where the proposed system exhibits a lower forgetting rate than TSCIL but falls short in terms of average accuracy; understanding the characteristics of this dataset and what makes it challenging for the proposed system, while developing specific strategies to improve performance without sacrificing stability, remains an unresolved issue. Evaluating the framework in truly online scenarios where class boundaries are not explicitly signalled, and where the model must detect and adapt to distribution shifts autonomously, would bring the system closer to real-world deployment conditions.

\section*{CRediT authorship contribution statement}
\noindent\textbf{Pablo García-Santaclara:} Conceptualization, Methodology, Software, Investigation, Writing -- original draft, Visualization.
\textbf{Bruno Fernández-Castro:} Conceptualization, Methodology, Investigation, Writing -- review \& editing, Visualization, Supervision, Funding acquisition.
\textbf{Rebeca P. Díaz-Redondo:} Conceptualization, Methodology, Investigation, Writing -- review \& editing, Visualization, Supervision, Funding acquisition.

\section*{Declaration of competing interests}
The authors declare that they have no known competing financial interests or personal relationships that could have appeared to influence the work reported in this paper.

\section*{Data availability}
The datasets used in this work are publicly available. Pre-processed versions were obtained from the TSCIL benchmark repository.

\section*{Acknowledgments}
This work was partially funded by the grant PID2023-148716OB-C31 funded by MCIU/AEI/10.13039/501100011033 (DISCOVERY project) and by the Galician Regional Government under project ED431B 2024/41 (GPC).

\bibliographystyle{plainnat}
\bibliography{cas-refs}

\clearpage
\FloatBarrier

\appendix
\section{Statistical Feature Definitions}
\label{app:features}
\renewcommand{\thesubsection}{\Alph{section}.\arabic{subsection}}
\noindent Let $x = [x_1, x_2, \dots, x_T]$ denote a single channel of a time series
sample of length $T$, and let $X = [X_1, X_2, \dots, X_{\lfloor T/2 \rfloor + 1}]$
denote its discrete Fourier transform (DFT) coefficients with corresponding
frequencies $f_k$, computed at sampling frequency $f_s$.

\vspace{6pt}
\noindent\begin{minipage}{\textwidth}

\captionof{table}{Formal definitions of the statistical features extracted per channel.}
\label{tab:feature_definitions}
\begin{tabular}{llp{7cm}}
\toprule
\textbf{Domain} & \textbf{Feature} & \textbf{Definition} \\
\midrule
\multirow{10}{*}{Time}
  & Mean               & $\mu = \frac{1}{T}\sum_{t=1}^{T} x_t$ \\[6pt]
  & Std.\ deviation    & $\sigma = \sqrt{\frac{1}{T}\sum_{t=1}^{T}(x_t - \mu)^2}$ \\[6pt]
  & Minimum            & $\min(x)$ \\[6pt]
  & Maximum            & $\max(x)$ \\[6pt]
  & Range              & $\max(x) - \min(x)$ \\[6pt]
  & MAD                & $\frac{1}{T}\sum_{t=1}^{T}|x_t - \mu|$ \\[6pt]
  & RMS                & $\sqrt{\frac{1}{T}\sum_{t=1}^{T} x_t^2}$ \\[6pt]
  & Skewness           & $\frac{1}{T\,\sigma^3}\sum_{t=1}^{T}(x_t-\mu)^3$ \\[6pt]
  & Kurtosis           & $\frac{1}{T\,\sigma^4}\sum_{t=1}^{T}(x_t-\mu)^4 - 3$ \\[6pt]
  & Zero-crossing rate & $\frac{1}{T}\sum_{t=1}^{T-1}\mathbf{1}[\tilde{x}_t \cdot \tilde{x}_{t+1} < 0]$ \\[6pt]
\midrule
\multirow{4}{*}{Frequency}
  & Spectral energy      & $\sum_{k} |X_k|^2$ \\[6pt]
  & Dominant frequency   & $f_k^* = \arg\max_k |X_k|^2$ \\[6pt]
  & Spectral centroid    & $f_c = \frac{\sum_{k} f_k\, |X_k|^2}{\sum_{k} |X_k|^2}$ \\[6pt]
  & Spectral spread      & $\sqrt{\frac{\sum_{k}(f_k - f_c)^2\, |X_k|^2}{\sum_{k}|X_k|^2}}$ \\[6pt]
\bottomrule
\end{tabular}
\end{minipage}

\vspace{4pt}
\noindent where $\tilde{x}_t = x_t$ if $x_t \neq 0$, else $\tilde{x}_t = 1$
(zero values are treated as positive to avoid ambiguity in sign changes).

\end{document}